# Comparing Methods for segmentation of Microcalcification Clusters in Digitized Mammograms

Hajar Moradmand[1], Saeed Setayeshi [2] and Hossein Khazaei Targhi[3]

[1] Islamic Azad University-Khorasgan Branch, Khorasgan, Iran

[2] Department of Biomedical Radiation Engineering, Amirkabir University of Technology
Tehran, Iran

[3] Islamic Azad University-Khorasgan Branch, Khorasgan, Iran

**Abstract**
The appearance of microcalcifications in mammograms is one of the early signs of breast cancer. So, early detection of microcalcification clusters (MCCs) in mammograms can be helpful for cancer diagnosis and better treatment of breast cancer. In this paper a computer method has been proposed to support radiologists in detection MCCs in digital mammography. First, in order to facilitate and improve the detection step, mammogram images have been enhanced with wavelet transformation and morphology operation. Then for segmentation of suspicious MCCs, two methods have been investigated. The considered methods are: adaptive threshold and watershed segmentation. Finally, the detected MCCs areas in different algorithms will be compared to find out which segmentation method is more appropriate for extracting MCCs in mammograms.

## 1. Introduction

Breast cancer is one of the most important factors of mortality in women over the world. In the United States alone, a most recent survey estimated 207,090 new cases of breast cancer and 39,840 deaths in women during 2010 [1]. Causes of this disease still remain unknown so; there is no sure way to prevent breast cancer. Early detection is the key to improve breast cancer treatment. Up to now, mammography remains the most effective diagnostic technique for early breast cancer detection. Radiologists is capable for detection the abnormalities of cancerous cells such as calcification, masses, architectural distortion, and asymmetry between breasts, breast edema and lymphadenopathy from mammogram images. Two major of mammographic abnormalities in breast cancer is calcifications and masses. Calcifications are small mineral deposits within the breast tissue, which look like small white spots on the films. They are often important and common findings on a mammogram. They can be produced from necrotic cellular debris or from cell secretion. They may be intramammary, within and around the ducts, within the lobules, in vascular structures, in interlobular connective tissue or fat. Alternatively, they may be found in the skin. Calcifications can appear with or without an associated lesion, and their morphologies and distribution provide clues as to their etiology as well as whether they can be associated with a benign or malignant process. There are two types of calcifications: macrocalcifications and microcalcifications. Accurate segmentation of individual microcalcifications is challenged by microcalcifications size and shape variability, superimposed surrounding tissues and high frequency noise [2]. Elder and Horsch a thorough review of the proposed methods for microcalcifications segmentation is provided [3]. Segmentation of individual microcalcifications has been achieved by grey-level based methods with empirically defined parameters such as region growing [4] and grey-level thresholding on pre-processed regions of interest (ROIs) [5]. To fulfill requirements for real-time behavior and parameter-free segmentation, more sophisticated techniques have been proposed such as morphologic operations [6], watershed algorithms [7], Bayesian pixel classification combined with Markov Random Field models [8] and radial gradient-based methods [9]. Furthermore, the wavelet transform has been used for the segmentation of microcalcifications, due to its ability to spatially localize high frequency components [10]. Recently, a segmentation method based on multiscale active rays was proposed to deal with microcalcification size variability [11]. Although the spectrum of computational methods has been proposed for detection of MCCs is wide, automated interpretation of microcalcifications still remains very difficult. This paper has been proposed two methods for MCCs segmentation. Although, these methods had been used but the way that they have been implemented in this paper is new and achieved very good segmentation result. The rest of paper is organized as follows. The proposed approach for segmentation of MCs with two methods: Morphology operation and watershed segmentation has been described in Section 2. The subsequent section discusses the obtained results (section3), and Section 4 presents the conclusions and future research directions.

## 2. The proposed method

This paper presented two methods for MCCs segmentation in digital mammograms. A block diagram of the proposed method is shown in Fig. 1.

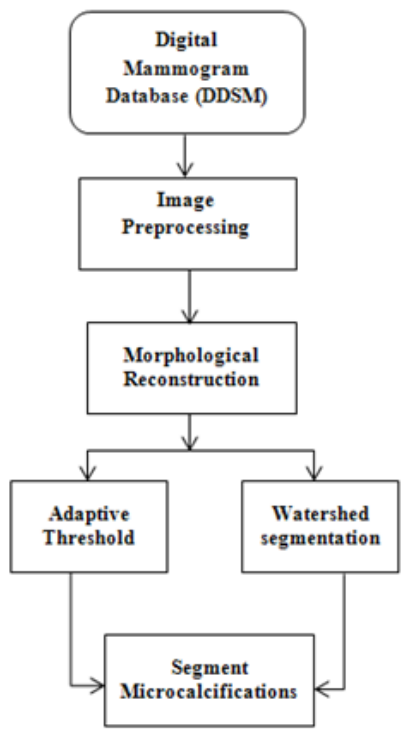

Fig. 1 Block diagram of the proposed method.

### 2.1 Databank

In this research Digital Database for Screening Mammography (DDSM) from University of South Florida is used for experiments. It was downloaded from http://marathon.csee.usf.edu/Mammography/DDSM.html.
Images containing suspicious areas have associated pixel-level ground truth information about the locations and types of suspicious regions. Also provided is software both for accessing the mammogram and truth images and for calculating performance figures for automated image analysis algorithms [12]. The use of such a database aids in comparison of CAD algorithms through evaluation using a common database. Our experimental data contains 126 images, from only craniocaudal (CC) view. The sampling rates of mammogram images are digits between 42 to 50 microns.

### 2.2 Preprocessing

A preprocessing step is performed in order to facilitate the subsequent MC segmentation task. In our algorithm this step includes three phases: masking, denoising, and image enhancement.

#### 2.2.1 Mask Generation

First, breast tissue is separated from the other parts of image which are completely dark. This allows processing of only the tissue region in further steps and eliminating the artifacts out of the mammogram. To accomplish this, a segmentation mask is used to separate the tissue region from the film region. The mask template is a binary matrix of size equal to that of the original image. Morphology operation and Otsu algorithm had been used to create appropriate mask.

#### 2.2.2 Denoising

Digital mammography images often contain significant amounts of noise which, accommodation with salt and pepper type noise. This noise should be removed for avoiding deterioration of the contrast enhancement algorithm and negative influence on the whole detection procedure. Therefore, de-noise the image with a 4×4 median filter. Median filter is a non-linear filter having the ability to remove salt and pepper noise without blur edges especially is a powerful tool for improving mammogram images.

#### 2.2.3 Image Enhancement

The contrast of a mammogram image is often poor, especially for dense, glandular tissues. In these cases distinguish MCs is quite hard. To overcome this obstacle contrast enhancement algorithm had been used. The aim of contrast enhancement is to increase the contrast of MC over the threshold. Our system performs automatic contrast enhancement, using a method based on unsharp masking and 2D discrete wavelet transform. To egregious edges and small details in the mammogram image Unsharp masking filter is very useful. The unsharp masking can be expressed as:

$$D_p(x,y) = D_o(x,y) + K(x,y) \times \left[D_o(x,y) - \frac{1}{mn}\sum_{j=1}^{n}\sum_{i=1}^{m} D_o(x_I, y_I)\right] \quad (1)$$

Where $D_o(x, y)$ and $D_p(x, y)$ are the densities of the original and the processed mammograms, respectively. The last term is the unsharp term with an $m \times n$ area centered at pixel $(x, y)$, $k(x, y)$ is a weight factor [2]. The mask size and the weight factor determine the frequency range and the degree of enhancement. The unsharp masking method reduces the low-frequency information while amplified the high-frequency detail. However, these processes could change the images dramatically to be applied to the mammograms.

Wavelet transform is a powerful tool for filtering that represents images hierarchically on the basis of scale and resolution, analyzing high-spatial frequency phenomena localized in space, and, thus can effectively extract information derived from localized high-frequency signals, such as those emitted by microcalcification. The two dimensional discrete wavelet transform (DWT) decomposes the approximation coefficients at level j into approximation coefficients at level j+1 and three sets of detailed coefficients, i.e., horizontal, vertical and diagonal Fig. 2 describes the basic decomposition step for image.

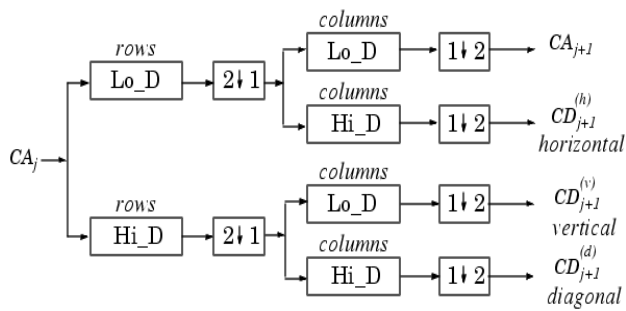

Fig. 2 Wavelet decomposition of an image. Two-dimensional wavelet transform leads to a decomposition of approximation coefficients at level j in four components: the approximation at level j+1, and the details in three orientations (horizontal, vertical, and diagonal).

The detailed coefficients contain small-scale components of the image. In frequency domain analysis, high frequency coefficients are detailed. Microcalcification often form on the mammogram image of fine grains appear bright in the breast tissue. So, we can assume that a wavelet decomposition of the mammogram image will contain the MC mostly within the detailed coefficients [10]. Five-level discrete wavelet decomposition had been employed by using Asymmetric Daubechies of order 8 since it accumulates more energy corresponding to the details of the wavelet transform and moreover it is characterized by symmetry and finite length to enhance mammograms. Due to these features, they can achieve high correlation with the clustered MC, and, therefore, they can effectively enhance MC. So, the filtered image is subjected to five-level discrete wavelet decomposition. This produces an approximation and five sets of horizontal, vertical and diagonal detailed coefficients. Afterwards, the contrast enhanced mammogram is obtained by reverse wavelet transform.

2.3 Segment Microcalcifications

For reasons such as; characteristics of breast tissue varies in texture and small size Microcalcification their detection and isolation of tissue still remains difficult. Select appropriate threshold is a sensible step, as a too high threshold can neglect the MCs which present less contrast, while a too low threshold makes that brilliant points which are selected do not correspond to some microcalcification; these points are caused by the oscillations in the grey levels of the background, due to the noise which contaminates the image. To overcome this problem first morphological reconstruction (opening and then closing by-reconstruction) had been used to clean up the image and smoothing so the foreground that contains breast tissue would be more recognizable. Because, comparing reconstruction-based opening and closing to standard opening and closing are more effective at removing small blemishes without affecting the overall shapes of the objects. Reconstruction is a morphological transformation involving two images and a structuring element (instead of a single image and structure element). One image, the marker, is the starting point for the transformation. The other image, the mask contains the transformation. The structure element used defined connectivity. If $g$ is the mask and $f$ is the marker, the reconstruction of $g$ from $f$, denoted; $R_g(f)$. The high points, or peaks, in the marker image specify where processing begins. The processing continues until the image values stop changing.

The following steps for morphologically reconstruct of mammogram images has been performed. First a marker image has been created. The characteristics of the marker image determine the processing performed in morphological reconstruction. The peaks in the marker image should identify the location of objects in the mask image. To create a marker image, first compute the background by imopen, and then subtract it from the mask image by imsubtract as indicated in Eq. (2) and Eq. (3).

$$\text{Background} = \text{imopen}(A, SE) \qquad (2)$$

$$\text{Marker} = \text{imsubtract}(A, \text{background}) \qquad (3)$$

Next the opening-by-reconstruction computed by imreconstruct as shown in Eq. (4). In the output image, all the intensity fluctuations except the intensity peak have been removed.

$$\text{Recon} = \text{imreconstruct}(\text{marker}, \text{mask}) \qquad (4)$$

Following the opening with a closing can remove the dark spots and stem marks, so try imclose followed by imreconstruct.

### 2.3.1 Adaptive Thresholding

Whereas, there are large variations threshold from one image to the next, so a constant threshold will not be good enough. To find a threshold for a particular image adaptive thresholding has been implemented. Local adaptive thresholding selects an individual threshold for each pixel based on the range of intensity values in its local neighborhood. This allows for thresholding of an image whose global intensity histogram doesn't contain distinctive peaks. The result is a binary image representing the suspicious areas that contains MCs. to achieved adaptive thresholding, first determined the image's histogram. As be illustrated in Fig. 3 the gray level of all histogram contain two Gaussian distribution function.

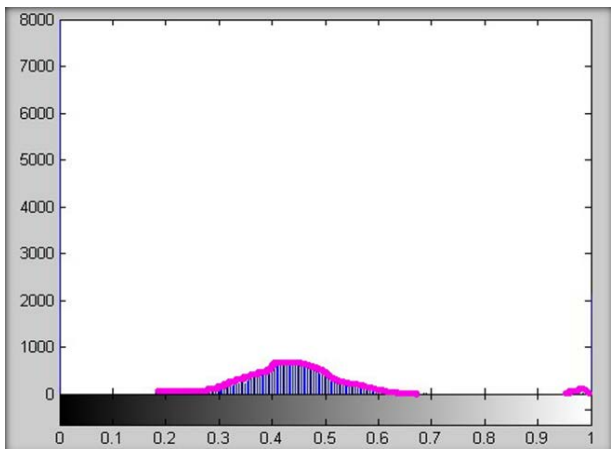

Fig. 3 An example of a mamogram's histogram.

Therefore, two gaussian functions have been adapted over histograms then mean and variance of them has been computed. In order to select a more accurate threshold Eq. (5) has been considered

$$Threshold = \frac{\frac{c_1}{\sigma_1} + \frac{c_2}{\sigma_2}}{\frac{1}{\sigma_1} + \frac{1}{\sigma_2}} \quad (5)$$

Where $c$ is the mean and $\sigma$ is the variance.

### 2.3.2 Watershed Segmentation

The watershed transform finds catchment basins and watershed ridge lines in an image by treating it as a surface where light pixels are high and dark pixels are low. The watershed transformation considers the gradient magnitude of an image as a topographic surface. Pixels having the highest gradient magnitude intensities (GMIs) correspond to watershed lines, which represent the region boundaries. Water placed on any pixel enclosed by a common watershed line flows downhill to a common local intensity minimum (LIM). Pixels draining to a common minimum form a catch basin, which represents a segment. Sobel filter is used often for approximates the maximum gradient of the image by giving more weight to the pixels nearest to $(i, j)$. It is equal to Eq. (6).

$$\sqrt{(\frac{\partial f_{ij}}{\partial x})^2 + (\frac{\partial f_{ij}}{\partial y})^2} \quad (6)$$

Direct application of the watershed segmentation to a gradient image can be lead to over-segmentation due to serious noise or image abnormality. Segmentation using the watershed transforms works better if, foreground objects and background would be marked whose goal is to detect the presence of homogeneous regions from the image. Internal markers are inside each of the objects of interest and external markers are contained within the back-ground. For segmentation with Marker-controlled watershed follow this basic procedure:

1. Compute foreground markers. These are connected blobs of pixels within each of the objects.

2. Compute background markers. These are pixels that are not part of any object.

3. Modify the segmentation function so that it only has minima at the foreground and background marker locations.

4. Compute the watershed transform of the modified segmentation function.

## 3. Results

The proposed algorithm operates on 126 mammograms image that contain MCCs from DDSM. The experimental result demonstrated that watershed segmentation is more accurate that adaptive thresholding but is more time consuming. Fig. 4 shows an example of our method on a mammogram image with adaptive thresholding method and in Fig. 5 with watershed segmentation.

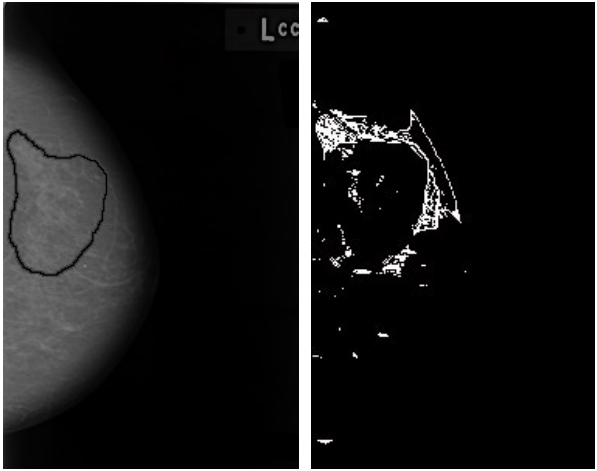

Fig. 4 Microcalcification segmentation with adaptive threshold.

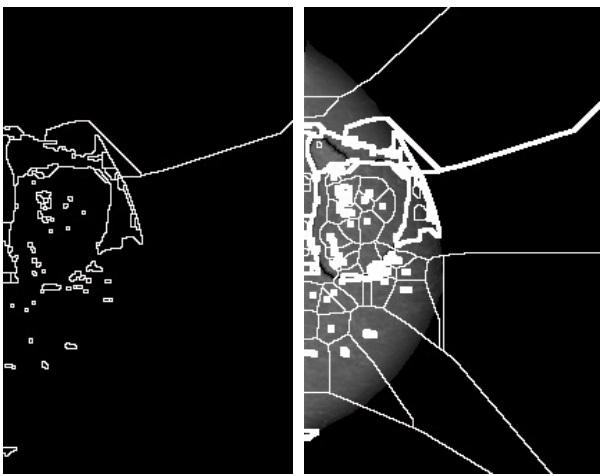

a) Watershed ridge lines    b) Markers and object boundaries superimposed on original image

Fig. 5 Microcalcification segmentation with watershed segmentation.

## 4. Conclusions

In this paper a computer system devised that support a radiologist in small field of digital mammography has been proposed. As noted earlier, MCs has very small and ubiquitous nature. So, we focused on the detection of clusters of microcalcifications. For this reason two method of segmentation had been implement on 126 mammograms. Result shows that watershed segmentation is more accurate than adaptive thresholding but more time consuming. All code required to accomplish these tasks was written in MATLAB. The system was evaluated by two radiologists. Their results show that the system introduces an improvement in the breast cancer detection. In our future research, we would like to evaluate the proposed method on more mammograms from clinical images and other database. We would also like to extend this research for detect and classification of other factors of breast cancer such as mass.

### Acknowledgments

The authors would like to thank Dr. Mohammad Esmail Akbari master of Iranian Cancer Research Center for his sincere contributions.